\documentclass[10pt]{IEEEtran}
\usepackage{amsmath,amsfonts}
\usepackage{algorithmic}
\usepackage{array}
\usepackage[caption=false,font=normalsize,labelfont=sf,textfont=sf]{subfig}
\usepackage{textcomp}
\usepackage{stfloats}
\usepackage{url}
\usepackage{verbatim}
\usepackage{graphicx}
\usepackage{balance}
\usepackage{tabularx}
\usepackage{booktabs}
\usepackage{multirow}
\usepackage{xcolor}
\usepackage{makecell}
\usepackage[numbers,sort&compress]{natbib}
\usepackage[table]{xcolor}
\usepackage{hyperref}
\usepackage{ulem}
\usepackage{subcaption}

\definecolor{ccr}{RGB}{0,0,255}  

\def\BibTeX{{\rm B\kern-.05em{\sc i\kern-.025em b}\kern-.08em
		T\kern-.1667em\lower.7ex\hbox{E}\kern-.125emX}}
	
\hypersetup{colorlinks=true,
	colorlinks=true,
	linkcolor=ccr,
	anchorcolor=ccr,
	citecolor=ccr}

\begin{document}
	\title{Bridging Modalities and Tasks: A Unified Hierarchical ViT for SAR-to-Optical Translation and Semantic Segmentation}
	\author{Siyuan Liu, Xuze Zhang, Yongshun Wang, Licong Pan, Hang Liu, and Huihui Li
		\thanks{Siyuan Liu, Licong Pan and Huihui Li are with the School of Automation, Northwestern Polytechnical University, Xi'an 710129, China (e-mail: lsy648@mail.nwpu.edu.cn, panlicong@mail.nwpu.edu.cn, lihhui@nwpu.edu.cn).}
		\thanks{Xuze Zhang, Yongshun Wang and Hang Liu are with the School of Cybersecurity, Northwestern Polytechnical University, Xi'an 710129, China (e-mail: zhangxuze@mail.nwpu.edu.cn, ys.wang@mail.nwpu.edu.cn, liuhang@nwpu.edu.cn).}
	}

	\maketitle
	
	\begin{abstract}
		
	Synthetic Aperture Radar (SAR) remains available through cloud and darkness, yet speckle and non-intuitive scattering make it hard to inspect and interpret. SAR-to-optical (S2O) translation renders radar observations as optical-like images for visual access, while semantic segmentation on SAR remains necessary since land-cover decisions cannot rely on synthesised pixels. We propose BMT, a hierarchical Vision Transformer coupling both tasks on a shared encoder via multi-scale one-hot conditioning, a gated LocalViTBlock, a colour-calibrating generation head, ControlNet-style zero-initialized injection, and uncertainty-based task weighting. The joint framework’s value is not label-free training: it uses paired optical images and the same pixel annotations as a dedicated SAR segmentor, supervising the SAR decoder and teacher-forcing the generator with ground-truth one-hot maps. At test time, the model takes only SAR and returns an optical-like readout and a radar-decided land-cover map, with the translation objective supplying dense appearance supervision that speckle lacks. On WHU-OPT-SAR, the predicted-condition model achieves the best SSIM and PSNR among compared translators; its land-cover map surpasses the same backbone trained for labelling alone and is within about one point of DeepLabV3+.
	\end{abstract}
	
	
	\section{Introduction}
	
	Synthetic Aperture Radar (SAR) provides all-weather, day-and-night imagery that remains available when optical acquisition fails. Speckle, geometric distortion and non-intuitive scattering nonetheless make SAR difficult to inspect by eye and difficult to label with models trained on optical imagery. Optical images are readily readable, but they drop out under cloud, haze and night---precisely when the radar observation is still there. A SAR-only segmentor still returns no inspectable image; a translator that only paints an optical-like picture still leaves land-cover unlabelled, or labelled on pixels that no optical sensor measured. Dedicated SAR segmentors already train with pixel annotations; the question is what else those annotations, together with paired optical images, can provide at test time, when only SAR remains.

	SAR-to-optical (S2O) translation and SAR semantic segmentation address those two gaps at the moment of use, yet they have largely been studied as separate tasks. On the translation side, generative adversarial networks (GANs)~\cite{mirza2014conditional,p2p,cycle} and, more recently, diffusion models~\cite{ho2020denoising,bib41,bib42} have produced increasingly realistic pseudo-optical images. On the segmentation side, convolutional networks, Transformers and hybrid architectures~\cite{segformer,hrnet,deeplabv3plus,segnext} have reached high accuracy on remote sensing benchmarks. Progress on each task does not supply the test-time product of the other. Photometric S2O may look plausible while field boundaries, water bodies and roads are distorted or invented, so the inspection product cannot be trusted as structure. Segmentors trained on SAR alone remain sensitive to speckle, so the interpretation product is starved of the appearance signal that optical models take for granted. Studies that couple the two usually do so sequentially, with a fixed task order~\cite{11322860}: translation first, then segmentation. Information then flows in only one direction. The second task cannot revise the pixels of the first, errors made early in the chain are inherited, and the optical reconstruction never becomes a training signal for the SAR encoder. The labelled paired set is then spent on two disjoint stages rather than on one model that, at test time, needs only SAR.

	The value of a joint framework is to spend a paired, fully labelled training set on a model that no longer needs that pair at test time---not to train with fewer radar labels than a dedicated segmentor. Training uses SAR tiles, registered optical images, and the same pixel maps a SAR-only decoder would require. The maps supervise segmentation and, as ground-truth one-hot conditions, teacher-force the generator; the optical images are the reconstruction target. That is more supervision at training time than a SAR-only segmentor, not less. What it buys is a deployable forward pass that takes only SAR: an optical-like readout the operator can inspect when Gaofen-1 is blocked, and a land-cover map decided on the radar measurement rather than on synthesised pixels. On the same annotated tiles, optical reconstruction further supplies a dense appearance target that speckle does not provide, so the SAR decoder is regularized by a readable signal rather than by backscatter alone. The translated image is therefore not a prettier SAR picture for its own sake, and the SAR map is not a by-product of generation. They are the two test-time products of one labelled, paired training run. At inference the ground-truth map is not used: segmentation runs first and its prediction conditions generation. The reported inspection scores are this predicted-condition setting; a ground-truth-condition run is an upper bound on how much the generator still depends on the quality of that predicted map.

	We propose BMT, a hierarchical Vision Transformer that instantiates this coupling rather than a serial pipeline. A shared encoder feeds two task-specific decoders, so both objectives shape the same multi-scale features: the generation decoder is the optical-like readout, and the segmentation decoder is the interpretation of the radar observation, trained with the appearance cue that SAR itself does not provide. The architecture follows the two halves of the value argument. On the translation side, a ControlNet-style path encodes the one-hot semantic map into a multi-scale pyramid and injects it at encoder stages through zero-initialized convolutions, inactive at initialization; a generation head then calibrates colour statistics and suppresses upsampling artefacts, so that the readout matches the optical domain in colour as well as in layout. On the interpretation side, a gated LocalViTBlock mixes global self-attention with spatial depthwise convolution, and uncertainty-based weighting learns how much each loss may move the shared representation, so that the optical-appearance signal can regularize SAR features without either task dominating.

	On WHU-OPT-SAR both outputs are evaluated under a scene-disjoint split. With generation conditioned on the predicted map, BMT leads the compared translators on SSIM and PSNR, while ControlNet is closer on FID and KID. The jointly trained decoder labels the SAR image itself. Segmentation is within about one point of DeepLabV3+ in mIoU, below recent dedicated models such as DOCNet, SegFormer and HRNet on thin objects, and higher than the same backbone trained for labelling alone on the same tiles. Neither a dedicated segmentor nor an unconditional translator returns both an optical-like image and a SAR-side map from SAR alone. Optical-to-SAR is discussed in Section~\ref{sec:s2o}. The main contributions are as follows.

\begin{enumerate}
	\item A joint S2O--segmentation framework trained with paired optical images and full pixel labels, then deployed on SAR alone: the labels supervise segmentation and teacher-force the generator; test time returns an optical-like image and a SAR-side map.
	
	\item ControlNet-style multi-scale injection of the one-hot category map through zero-initialized convolutions, plus a generation head for colour calibration and anti-aliasing. Ground-truth maps are used only in training; reported generation scores use the predicted map.
	
	\item A gated LocalViTBlock and uncertainty-based weighting on the shared encoder. Most of the labelling gain over a SAR-only head comes from the translation objective; these two blocks still add smaller additional gains.
\end{enumerate}

	\section{RELATED WORK}
	
	\subsection{Remote Sensing Image Semantic Segmentation}
	
	Semantic segmentation of remote sensing imagery aims at pixel-wise classification of ground objects and is widely adopted for land-use mapping, urban monitoring, and disaster assessment. Under deep learning, early work relied on encoder-decoder architectures. U-Net~\cite{unet} fuses multi-level features through skip connections to recover spatial details. To capture multi-scale context, the DeepLab family~\cite{deeplabv3plus} introduced atrous convolution and atrous spatial pyramid pooling (ASPP). PSPNet~\cite{pspnet} designed a pyramid pooling module to aggregate global contextual information, while HRNet~\cite{hrnet} maintains high-resolution streams in parallel with multi-scale fusion, improving boundary localization. Despite these advances, convolutional networks remain limited in modeling long-range dependencies in large-scale remote sensing scenes.
	
	Transformer architectures have enhanced global modeling through self-attention. SegFormer~\cite{segformer} adopts a hierarchical Transformer encoder with a lightweight multilayer perceptron (MLP) decoder, achieving an excellent accuracy-efficiency trade-off. Swin Transformer~\cite{swin} leverages window-based self-attention with shifted windows for linear complexity. SegNeXt~\cite{segnext} re-examines convolutional attention via a multi-scale convolutional attention module. More recently, hybrid architectures combining CNNs and Transformers~\cite{huang2024survey} have demonstrated the benefits of balancing local details and global semantics. Class-aware and frequency-domain networks such as LoG-CAN~\cite{logcan}, DOCNet~\cite{docnet}, SFFNet~\cite{sffnet}, AFENet~\cite{afenet} and DeepKANSeg~\cite{deepkanseg} further raise single-task accuracy on remote sensing benchmarks; they are compared as dedicated SAR segmentors in Section~IV. All of these designs remain single-task oriented and do not accommodate the joint optimization of translation and segmentation.
	
\subsection{SAR-to-Optical Image Translation}
\label{sec:s2o}
	
	Early S2O pipelines mapped SAR to optical colour with CNNs~\cite{wang2018generating}, but the outputs still differed from real optical imagery; SAR-GAN adopts a denoising-and-coloring cascade whose accuracy is limited by the fidelity of synthetic noise modeling. Conditional GANs~\cite{mirza2014conditional} are widely used due to their strong generative capability: Merkle et al.~\cite{merkle2018exploring} applied cGANs to SAR-optical matching, and BicycleGAN~\cite{bib25} injects noise into a cGAN to capture multimodal outputs. Cycle-consistency-based methods~\cite{cycle} further enabled unpaired translation by jointly training bidirectional networks. Several studies~\cite{bib10,bib11,bib17,bib35} focus on core challenges in S2O translation, while others~\cite{bib15} explore the impact on downstream classification.
	
	Diffusion models~\cite{ho2020denoising,bib41} offer an alternative via iterative denoising and have been used in Earth observation~\cite{fulvio_2023} and conditional S2O translation~\cite{10330015,Shi2024ABA}. ControlNet~\cite{zhang2023adding} conditions a generator with spatially aligned maps such as edge, depth or pose. Inspired by this, we inject the one-hot semantic map at multiple encoder levels through zero-initialized convolutions---a ControlNet-style design that, to our knowledge, has not been used for joint S2O translation and segmentation.

	This paper addresses the S2O direction because the target task is SAR interpretation: translation improves visual readability, the semantic map regularizes generation, and the translation objective in turn enriches SAR features. Optical-to-SAR (O2S) serves a different purpose---reproducing sensor-specific scattering and speckle---and requires evaluation criteria that are not comparable with the optical-domain metrics used here~\cite{wang2018generating,merkle2018exploring,10330015,Shi2024ABA}. The injection itself can take any spatially aligned map; encoding speckle statistics for O2S is a separate problem, which we leave to future work.
	
	\section{METHOD}
	
	\subsection{Overview}
	
	\begin{figure*}
		\centerline{\includegraphics[width=\linewidth,height=0.5\linewidth]{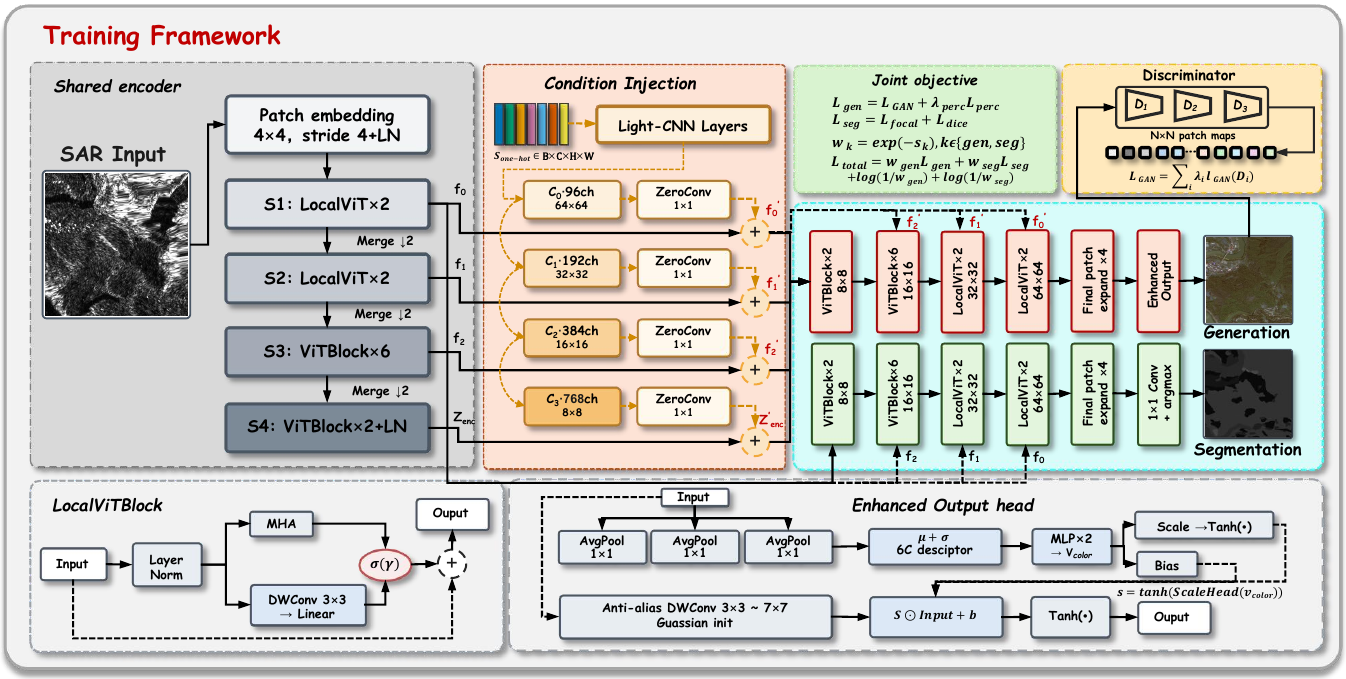}}
		\caption{Training pipeline of BMT. A four-stage hierarchical ViT encoder is shared by a generation decoder (with multi-scale color correction) and a segmentation decoder. The ground-truth one-hot semantic map is encoded into a multi-scale feature pyramid and injected at every encoder stage through zero-initialized convolutions. Uncertainty weights balance the two task losses during end-to-end training.}
		\label{fig1}
	\end{figure*} 
	
	\begin{figure*}
		\centerline{\includegraphics[width=\linewidth]{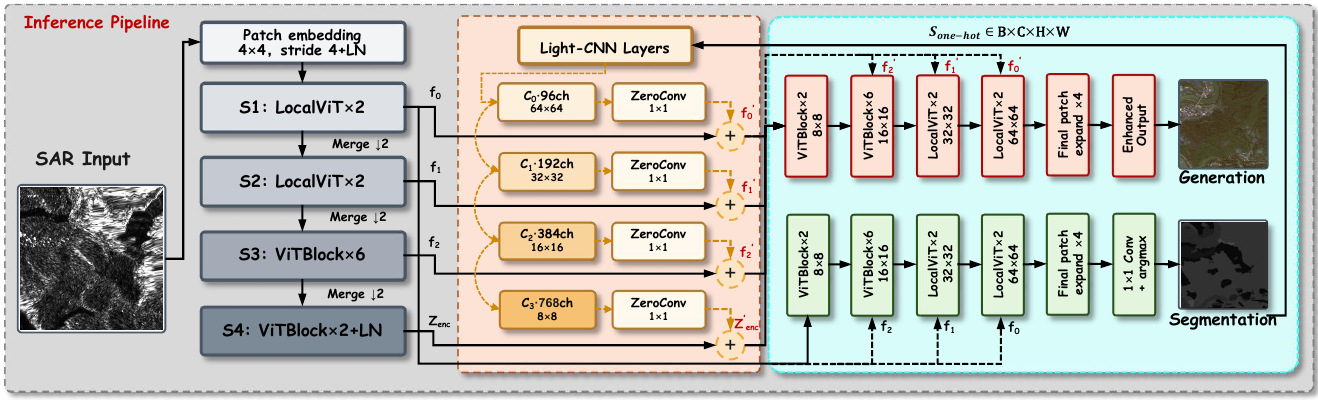}}
		\caption{Inference pipeline of BMT. The encoder and the condition injection follow Figure~\ref{fig1}. The segmentation branch first predicts a label map, which is converted to a one-hot condition for the generation pass, and no ground-truth annotation is used.}
		\label{fig:infer}
	\end{figure*}
	
	Figure~\ref{fig1} shows the training pipeline: a shared hierarchical ViT encoder, generation and segmentation decoders, and ControlNet-style multi-scale injection of the one-hot semantic map. Uncertainty weights balance the two task losses. During training, the ground-truth one-hot map is injected into the generation pass only; the segmentation branch sees only the SAR image. At inference (Figure~\ref{fig:infer}), segmentation runs first and its prediction conditions generation, so no annotation is used. The condition encoder, zero-initialized convolutions, both decoders and the discriminator are trained end-to-end. The generator follows a Pix2pix-style paired architecture.
	
	\subsection{Cooperative Hierarchical Encoding Strategy}
	
	In dense prediction tasks such as translation and segmentation, capturing fine-grained local details and long-range semantic dependencies is essential. Therefore, our shared encoder is designed as a collaborative multi-stage hierarchical architecture. This pyramid structure gradually reduces the spatial resolution while increasing the feature dimension, producing multi-scale representations that the decoder needs in order to recover accurate object boundaries.
	
	As shown in Figure~\ref{fig1}, given an input SAR image $\mathbf{x} \in \mathbb{R}^{3 \times H \times W}$, it is first projected into a sequence of non-overlapping patches through a convolutional layer. Between different stages, the Merging layer performs spatial downsampling, which mimics the inductive bias of convolutional neural networks and provides a computationally affordable way to process high-resolution features in the early stages.
	
	A core design choice is that the form of the Transformer block should adapt to the resolution of the current stage. In the shallow stages, the feature map still retains a relatively high spatial resolution. Using global multi-head self-attention in these stages not only incurs substantial computational overhead, but also tends to ignore local textures (edges, corners, fine structures) that are critical for segmentation. To address this trade-off, we design the LocalViTBlock, which performs global context aggregation and local pattern extraction in a single module. The LocalViTBlock operates through two parallel branches after layer normalization. The global branch performs standard multi-head self-attention (MHA) to capture long-range dependencies across the entire feature map; the local branch injects spatial locality through a depth-wise separable convolution (DWConv) followed by a linear projection. These two complementary representations are then fused via a learnable gating mechanism:
	
	\begin{align}
		\mathbf{x}_{\text{fused}} &= \sigma(\gamma) \odot \mathbf{x}_{\text{global}} + (1 - \sigma(\gamma)) \odot \mathbf{x}_{\text{local}}
	\end{align}
	
	where $\gamma$ is a learnable scalar, and $\sigma$ denotes the sigmoid function. This gate is initialized at $0$, so that $\sigma(\gamma)=0.5$ and the two branches contribute equally at the beginning of training. This balanced start stabilizes optimization and still allows the network to learn a stage- and task-dependent mixing ratio.
	
	At the deeper stage, the spatial resolution has been reduced by the previous Merging layer. At this point, the computation of global self-attention becomes affordable, and the features are increasingly semantic and abstract; local texture patterns have already been encoded in the shallow layers and transmitted via skip connections, so explicit local branches are no longer required. In these stages, we adopt the standard ViTBlock, relying entirely on global multi-head self-attention. This design simplifies the deep structure, allowing the model to focus its capacity on high-level semantic reasoning. 
	
	To enhance the representation capability at coarser scales, the deep stages contain more Transformer blocks than the shallow stages. In addition, the encoder stores the skip connection features from each stage, providing the decoder with rich multi-scale boundary and content cues to restore fine spatial details, and finally generates bottleneck features through layer normalization to be passed to the subsequent decoding branch.

	\subsection{Generation Branch}
	
	As shown in Figure~\ref{fig1}, the generator decoder branch starts from the bottleneck features $\mathbf{z}_{\text{enc}}$ and reconstructs a full-resolution RGB image through an upsampling architecture that is symmetric to the encoder and uses skip connections. The symmetric design allows the multi-scale features extracted by the encoder to be reused step by step, while the skip connections feed features from each stage of the encoder directly into the decoder layers of the corresponding resolution, providing high-fidelity positional and textural cues for restoring object boundaries and spatial details.
	
	Each decoder layer sequentially performs an Expand layer for spatial upsampling, fusion with skip features from the corresponding encoder stage, and a set of ViTBlock and LocalViTBlock transformations that are symmetric to those in the encoder. Among these, the LocalViTBlock is placed at the later, high-resolution decoder stages. At these stages the feature maps have large spatial dimensions and semantics become more concrete; global self-attention then struggles to restore subtle local textures, whereas the local inductive bias of depthwise convolution can suppress blurring, sharpen edges, and complement global attention. After passing through the final decoder layer, the FinalPatchExpand layer further increases the resolution, yielding a feature map of shape $B \times C_{\text{out}} \times H \times W$, which provides sufficient detail capacity for target image reconstruction.
	
	\subsubsection{Output Head}
	
	To produce a high-quality RGB output, the output head performs color correction by extracting color statistics at global, regional and local scales through multi-scale feature fusion, which provides comprehensive color priors. This module uses three parallel adaptive average pooling branches with output sizes of $1 \times 1$, $4 \times 4$, and $16 \times 16$, respectively, to obtain multi-scale feature maps. The mean and standard deviation are calculated separately for features at each scale, resulting in a total of $6 \times C_{\text{in}}$ statistical descriptors. These descriptors capture color distribution information ranging from overall hue to local texture variations. After being fused through a two-layer MLP, they generate a compact color condition vector:
	
	\begin{equation}
		\mathbf{v}_{\text{color}} = \text{MLP}\big([\mu_1, \sigma_1, \mu_4, \sigma_4, \mu_{16}, \sigma_{16}]\big)
	\end{equation}
	
	where $\mu_s$ and $\sigma_s$ are the mean and standard deviation at scale $s$. The color-conditioning vector is combined with locally transformed features and fed to scale and bias heads. The final output is:
	
	\begin{align}
		 &\mathbf{s} = \tanh(\text{ScaleHead}(\mathbf{f}_{\text{fused}})) \\
		 &\mathbf{b} = \text{BiasHead}(\mathbf{f}_{\text{fused}}) \\
		 &\mathbf{x}_{\text{out}} = \tanh\bigl( \mathbf{s} \odot \mathbf{x}_{\text{main}} + \mathbf{b} \bigr)
	\end{align}

	where $\mathbf{x}_{\text{main}}$ denotes the locally transformed features $\mathbf{f}_{\text{fused}}$ after the anti-aliasing convolution. The output head applies learned scale and bias modulation, and the statistics-based conditioning provides guidance for matching the color statistics of the target domain. The anti-aliasing convolution uses a depthwise convolution kernel with Gaussian-style initialization, which suppresses checkerboard artifacts from the upsampling decoder.

	\subsubsection{Multi-Scale Discriminator}

	To balance global structural consistency and local texture details, we adopt a multi-scale PatchGAN discriminator $\{D_1, D_2, D_3\}$, following the common practice in conditional image translation~\cite{wang2018high}, where each sub-discriminator $D_i$ outputs an $N \times N$ probability map, and a weighted adversarial loss is ultimately optimized, encouraging the generator to both preserve high-frequency texture details and maintain large-scale geometric consistency.

	\subsection{ControlNet-Style Condition Injection Module}
	
	The generation pass is conditioned on a spatially aligned one-hot map rather than on a global class vector. During training that map is the ground-truth one-hot (teacher forcing); at inference it is the predicted map, and no annotation is used. A global vector can shift overall appearance, but it does not specify where each land-cover category should appear. Following ControlNet~\cite{zhang2023adding}, we encode the one-hot map into a feature pyramid matched to the four encoder stages and add each pyramid level to the corresponding encoder feature through a zero-initialized convolution. Category layout is thereby available from the fine skip features to the bottleneck, while the conditional path starts as an identity mapping.
	
	\subsubsection{Condition Encoder}
	
	The control condition encoder takes the one-hot encoded segmentation map as its input. Given a segmentation label map $\mathbf{S} \in \mathbb{R}^{B \times H \times W}$ with $K$ categories, it is one-hot encoded into $\mathbf{S}_{\text{oh}} \in \mathbb{R}^{B \times K \times H \times W}$, which forms the condition:
	
	\begin{equation}
		\mathbf{C} = \mathbf{S}_{\text{oh}} \in \mathbb{R}^{B \times K \times H \times W}
	\end{equation}
	
	The condition is processed by a multi-stage lightweight hierarchical convolutional neural network encoder. Each stage includes a stride-2 convolutional layer, and the encoder produces a multi-scale feature pyramid whose spatial resolution and channel dimensions match those of each stage of the vision Transformer encoder. The condition can therefore be added at the same scale as the encoder feature it is meant to modulate.

	\subsubsection{Zero-Initialized Convolution Injection}
	
	Following the zero-initialized residual injection of ControlNet~\cite{zhang2023adding}, each conditional feature is added to the corresponding encoder feature through a zero-initialized 1×1 convolution. At initialization, the output of the zero-initialized convolution is zero, so the injection reduces to an identity mapping and the encoder starts from its unconditional behavior. As training progresses, the convolution weights grow from zero and the encoder can use the condition as a residual. Unlike the original ControlNet, which attaches its conditioning branch to a frozen copy of a pre-trained backbone, here the branch is injected into the shared encoder and the whole network is trained from scratch.
	
	\begin{align}
		\mathbf{f}_i' &= \mathbf{f}_i + \text{ZeroConv}(\mathbf{c}_i) \quad \text{for } i = 0, 1, 2 \\
		\mathbf{z}_{\text{enc}}' &= \mathbf{z}_{\text{enc}} + \text{ZeroConv}(\mathbf{c}_3)
	\end{align}
	
	where $\mathbf{f}_i$ are the encoder skip features at stages 0--2, and $\mathbf{z}_{\text{enc}}$ is the bottleneck feature. Features that are not consumed by any decoder path are deliberately left without injection.

	\subsection{Segmentation Branch Decoder}
	
	The segmentation decoder shares the same encoder and skip connections as the generative decoder, but is optimized for pixel-level classification. It adopts the same ViT decoder stage architecture as the generative decoder, using LocalViTBlocks at the later, high-resolution stages. It concatenates the skip features and processes them through a decoder stage containing ViTBlocks, and a $1\times1$ convolution produces pixel-level logits. The number of output channels equals the number of classes $K$ of the target dataset, so the same decoder applies to binary and multi-class problems alike.
	
	\subsection{Objective Function}

	\subsubsection{Adversarial Loss}
	The framework adopts a least-squares generative adversarial scheme. The discriminator and generator objectives are:
	
	\begin{align}
		\mathcal{L}_{D}^{gan} &= \mathbb{E}_{\mathbf{Y}_{gt}}\left[(D(\mathbf{Y}_{gt})-1)^2\right] + \mathbb{E}_{\mathbf{Y}_{gen}}\left[D(\mathbf{Y}_{gen})^2\right] \\
		\mathcal{L}_{gan} &= \mathbb{E}_{\mathbf{Y}_{gen}}\left[(D(\mathbf{Y}_{gen})-1)^2\right] \label{eq:gan}
	\end{align}
	
	\subsubsection{Perceptual Loss}
	
	To further improve perceptual quality, we introduce a multi-scale VGG-16 perceptual loss that promotes photorealism:
	\begin{equation}
		L_{\text{perc}}(G) = \sum_{\ell \in \mathcal{L}} w_\ell \cdot \big\|\phi_\ell(G(x)) - \phi_\ell(y)\big\|_1
	\end{equation}
	where $\phi_\ell$ denotes VGG-16 features after ReLU at layer $\ell$, and $\mathcal{L} = \{\ell_{1},\ell_{2},\ell_{3},\ell_{4}\}$ with corresponding weights $w = \{1.0, 1.0, 0.5, 0.25\}$.
	
	The total generation loss combines the adversarial term in Eq.~\eqref{eq:gan} with the perceptual term:

\begin{equation}
	L_{\text{gen}} = \mathcal{L}_{gan} + \lambda_{\text{perc}}L_{\text{perc}}
\end{equation}
	
	\subsubsection{Segmentation Loss}
	The segmentation loss combines Focal loss and Dice loss:
	
	\begin{equation}
		\begin{aligned}
			L_{\text{seg}}(\hat{y}, y) = 
			& -\frac{1}{N}\sum_{i=1}^{N} \omega_{y_i} (1 - \hat{y}_{i, y_i})^\gamma \log \hat{y}_{i, y_i} \\
			&+ \lambda \left(1 - \frac{2\sum_{i=1}^{N} y_{i,s} \hat{y}_{i,s} + \epsilon}{\sum_{i=1}^{N} y_{i,s} + \sum_{i=1}^{N} \hat{y}_{i,s} + \epsilon}\right)
		\end{aligned}
	\end{equation}
	
	where $\hat{y}_{i, y_i}$ is the predicted probability for the true class $y_i$, $\omega$ is a per-class weighting vector that compensates class imbalance, $\gamma \ge 0$ is the focal focusing parameter, and the Dice term is computed for each class $s$ and averaged over the $K$ classes.
	
	\subsubsection{Uncertainty-Based Multi-Task Weighting}
	Following Kendall et al.~\cite{kendall}, we formulate multi-task learning as uncertainty minimization. Each of the two tasks receives an uncertainty weight:
	
	\begin{equation}
		w_k = \exp(-s_k)
	\end{equation}
	
	where $s_k$ is the learnable log-variance. The total generator loss becomes:
	
	\begin{align}
		L_G^{\text{weighted}} &= w_{\text{gen}} \cdot L_{\text{gen}} + \log w_{\text{gen}}^{-1} \\
		L_S^{\text{weighted}} &= w_{\text{seg}} \cdot L_{\text{seg}} + \log w_{\text{seg}}^{-1} \\
		L_{\text{G-total}} &= L_G^{\text{weighted}} + L_S^{\text{weighted}}
	\end{align}

	The losses above fall into two groups. Adversarial and perceptual terms shape the appearance of the translated image and are combined with a fixed coefficient $\lambda_{\text{perc}}=0.2$. The segmentation loss supervises pixel-level labels on the same encoder. How much the shared representation should serve translation versus labeling is not a second hand-tuned coefficient: $w_{\text{gen}}$ and $w_{\text{seg}}$ are learned with the network, so the relative emphasis of the two tasks can change as training proceeds.
	
	\section{EXPERIMENTS}
	
	\subsection{Experimental Settings}
	
	\subsubsection{Dataset}
	
	We use the public WHU-OPT-SAR dataset~\cite{li2022mcanet}. WHU-OPT-SAR is a SAR--optical land-cover dataset of 100 fully registered image pairs (3704 $\times$ 5556 pixels). The SAR images are from Gaofen-3 and the optical images from Gaofen-1; both have a spatial resolution of 5~m. Pixel-wise land-cover annotations are provided with the imagery. We select five large scenes with diverse land-cover types. Each scene is tiled into $256\times256$ patches on a 224 pixel grid, discarding incomplete edge tiles, which produces 1920 candidate SAR--optical--label triplets. Because a 224 pixel stride leaves a 32 pixel overlap between adjacent tiles, overlapping patches are then discarded until the retained tiles share no geographic region. Patches whose machine-generated labels are unreliable are removed by manual inspection. This procedure yields 1098 spatially non-overlapping triplets, which are partitioned into 900 training and 198 test samples at the scene level: no parent scene contributes tiles to both splits, so the two sets share neither overlapping pixels nor a common geographic scene.
	
\subsubsection{Baselines}

For the generation task, we use CycleGAN~\cite{cycle}, Pix2pix~\cite{p2p}, Pix2pixHD~\cite{wang2018high}, S-CycleGAN~\cite{wang2019scyclegan}, StegoGAN~\cite{10656169}, PatchGCL~\cite{pcl}, WFLM-GAN~\cite{li2022multiscale}, HVTC-GAN~\cite{11322860} and ControlNet~\cite{zhang2023adding} as baselines on the WHU-OPT-SAR dataset. WFLM-GAN translates in two stages (wavelet grayscale, then colorization) without a semantic map; HVTC-GAN also generates unconditionally and defers segmentation to a later serial stage. ControlNet is a conditional generator with spatially aligned maps; unlike BMT it does not jointly train a SAR labelling head, so its train- and test-time condition is the ground-truth one-hot map rather than a predicted one. Its FID/KID scores are therefore not protocol-matched to BMT(pred). For the segmentation evaluation we use the widely adopted FCN-32s~\cite{long2015fully} as a fixed evaluator of the generated images, in addition to the segmentation decoder that is trained jointly within our framework. All models use publicly available implementations. For the segmentation comparison we additionally train FCN~\cite{long2015fully}, PSPNet~\cite{pspnet}, DeepLabV3+~\cite{deeplabv3plus}, U-Net~\cite{unet}, HRNet~\cite{hrnet}, SegFormer~\cite{segformer} and Swin-Unet~\cite{swinunet}, together with the recent remote sensing segmentors LoG-CAN~\cite{logcan}, SFFNet~\cite{sffnet}, DOCNet~\cite{docnet}, AFENet~\cite{afenet} and DeepKANSeg~\cite{deepkanseg}, on the same training split, with their official implementations, the same input resolution and a comparable training budget, and we evaluate all of them with the same metric suite. Methods whose interface is prompt- or vocabulary-driven are not included, because their evaluation protocol and label space differ from the closed-set seven-class setting adopted here, and the dataset used in this work is not designed to support that protocol.

\subsubsection{Evaluation Metrics}
To evaluate the translation results, we use four metrics: SSIM~\cite{wang2004image}, PSNR, Fréchet Inception Distance (FID)~\cite{heusel2017gans}, and Kernel Inception Distance (KID)~\cite{binkowski2018demystifying}. SSIM measures structural similarity between images, while PSNR evaluates pixel-level differences; higher values indicate closer agreement with the optical reference. FID and KID measure distributional differences in the image feature space; lower values indicate more realistic pseudo-optical images. For segmentation we report mean Intersection over Union (mIoU) and mean Pixel Accuracy (mPA): mIoU is the mean over classes of the overlap between prediction and ground truth, and mPA is the mean over classes of the fraction of correctly predicted pixels.

\begin{figure*}[h]
	\centering
	\includegraphics[width=\linewidth]{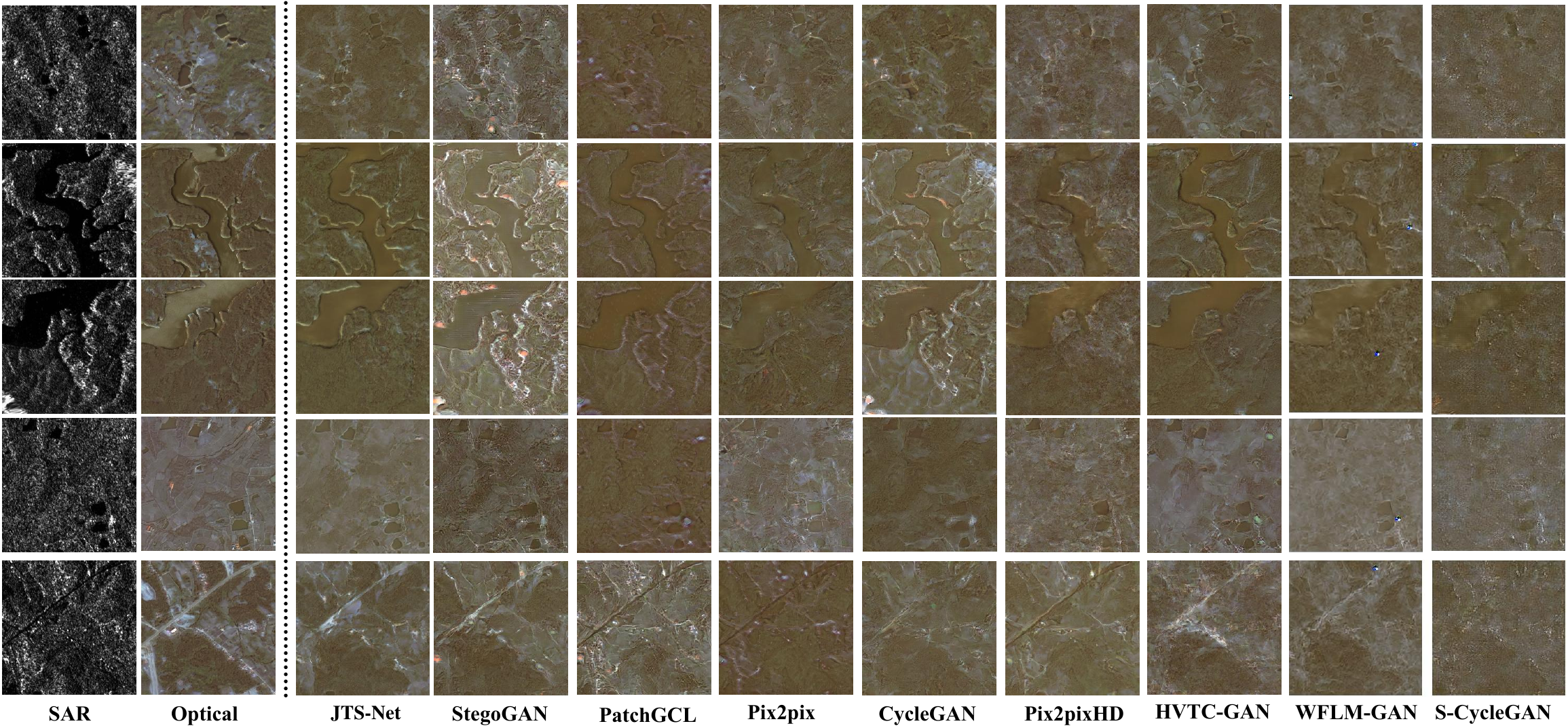}
	\caption{Visual comparison of S2O translation on WHU-OPT-SAR. Each row is one test patch. Columns show the input SAR image, the outputs of the compared translators, and the optical reference. BMT is shown under the predicted-condition protocol.}
	\label{fig:gen-whu}
\end{figure*}

\begin{figure*}[h]
	\centering
	\includegraphics[width=\linewidth]{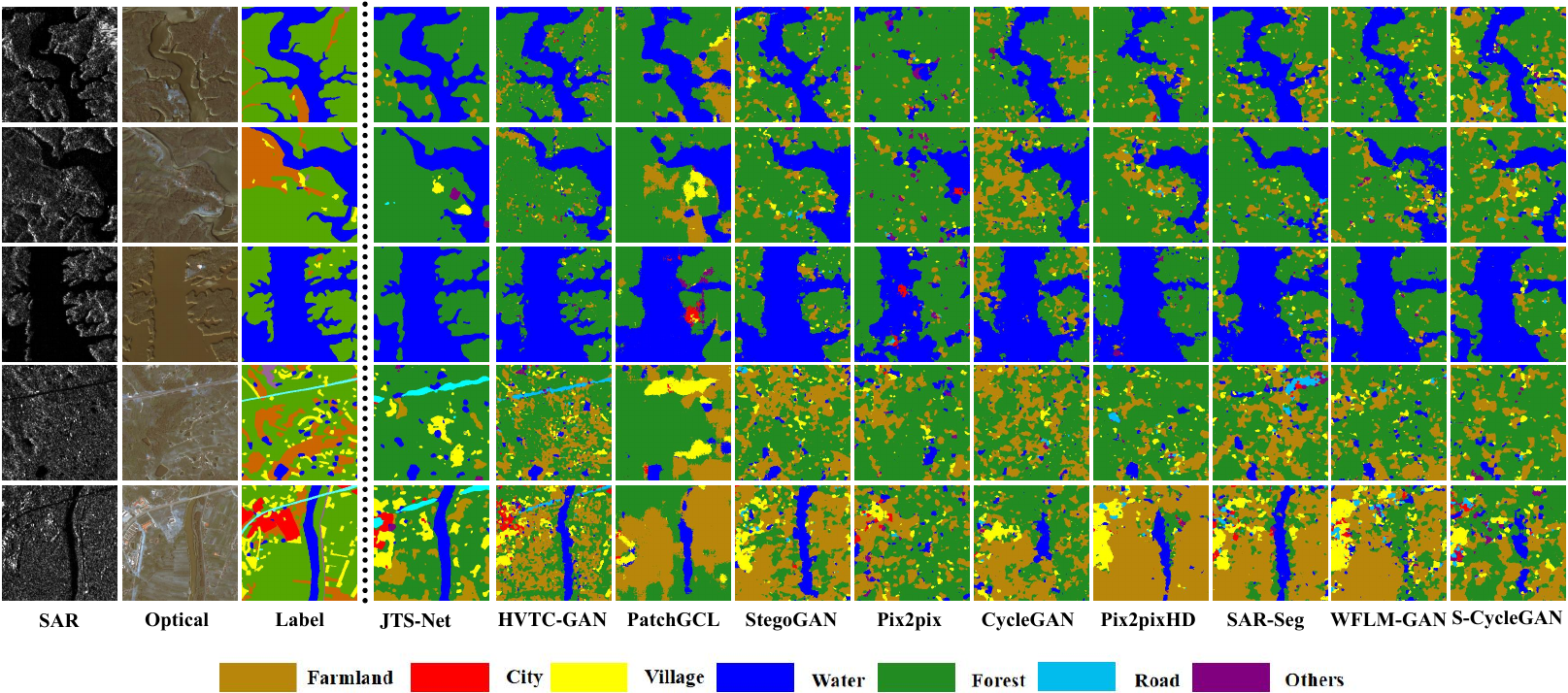}
	\caption{Land-cover maps obtained by applying the same pre-trained FCN-32s to the translated images on WHU-OPT-SAR. The segmentor is held fixed, so differences among columns come from the generated image rather than from a change of decoder.}
	\label{fig:seg-whu}
\end{figure*}

\subsubsection{Setup}
We train all models from scratch using the Adam optimizer with $(\beta_1, \beta_2) = (0.5, 0.999)$. The initial learning rates of the generator and the discriminator are both $1 \times 10^{-4}$ and decay to $1 \times 10^{-6}$. A linear-step schedule keeps the learning rate constant for the first $100$ epochs and linearly decays it over the subsequent $100$ epochs. The shared encoder learning rate is initialized at $0.5\times$ the decoder rate and annealed to $0.1\times$ between epochs $100$ and $200$, preventing late-stage segmentation gradients from dominating the shared representations. Three protocols are used. \emph{Predicted-condition} (default, all main results): the segmentation branch runs first and its prediction conditions the generator, so no annotation is used at test time. \emph{Ground-truth-condition}: the annotation conditions the generator and is reported only as an upper bound. \emph{No-condition}: the generator receives no semantic input; it appears only in the conditioning ablation.

\subsection{Results of WHU-OPT-SAR}
\label{sec:whu-results}

WHU-OPT-SAR is the paired test of the value claim in Section~I. Training uses registered optical images and full pixel labels; testing uses SAR alone under a scene-disjoint split. BMT is read here as one framework with two test-time products, not as a translator that happens to carry a segmentor, and not as a method that saves labels relative to a dedicated SAR decoder. The optical-like image is the inspection half when Gaofen-1 is absent and no annotation is used at test time. The SAR-side map is the interpretation half, regularized by the optical reconstruction target on the same labelled tiles. Neither half substitutes for the other.

Figure~\ref{fig:gen-whu} is the inspection half. The water scenes in the middle rows make the differences easiest to see. BMT keeps the water body in place and the shoreline comparatively sharp, whereas Pix2pix and Pix2pixHD leave a smeared edge. WFLM-GAN recovers coarse structure through its two-stage pipeline---wavelet fusion to a grayscale image, then colorization---but the colorized shoreline and surrounding hues still drift from the optical reference. HVTC-GAN is closer in overall tone, yet its generator is also unconditioned: the serial segmentation stage, which fuses a segmentor on the translated image with a pretrained SAR segmentor, does not constrain the pixels shown here. Unpaired CycleGAN and StegoGAN preserve high-contrast edges, but they invent land cover (spurious vegetation is common) and shift the color relative to the optical image. On the road row, BMT is the closest in both hue and boundary continuity. What the figure asks is whether the optical-like view can be inspected as structure: water, shoreline and road remain in place rather than being invented or washed out. That is the translation half of the coupling---a layout-constrained readout, not unconstrained photometric synthesis.

\begin{table}[htbp]
	\centering
	\caption{Comparison of SAR-to-optical translation on WHU-OPT-SAR. BMT(pred) is the deployable setting and uses the predicted-condition protocol, in which the semantic condition is produced by the model's own segmentation branch. BMT(gt) conditions on the ground-truth map and is reported only as an upper bound.}
	\renewcommand{\arraystretch}{1.0}
	\setlength{\tabcolsep}{3pt}
	\begin{tabular}{cc cccc}
		\toprule
		\textbf{Method} & \textbf{Year} & \textbf{SSIM} \(\uparrow\)(\%) & \textbf{PSNR} \(\uparrow\)(dB) & \textbf{FID} \(\downarrow\) & \textbf{KID} \(\downarrow\)(\(\times 100\)) \\
		\midrule
		CycleGAN  & 2017 & 39.63 & 19.82 & 174.18 & 7.72  \\
		Pix2pix   & 2017  & 39.25 & 21.54 & 192.21 & 10.98 \\
		Pix2pixHD & 2018  & 38.89 & 22.17 & 229.32 & 16.54 \\
		StegoGAN  & 2024  & 23.98 & 17.47 & 203.19 & 10.28 \\
		PatchGCL  & 2024  & 37.91 & 19.73 & 190.72 & 10.54 \\
		S-CycleGAN & 2019 & 35.21 & 21.96 & 311.45 & 28.15 \\
		ControlNet & 2023 & 36.24 & 21.66 & \uline{138.69} & \uline{3.98} \\
		WFLM-GAN   & 2022 & 43.17 & 22.04 & 251.40 & 17.19 \\
		HVTC-GAN   & 2026 & 43.54 & 22.18 & 171.11 & 6.34  \\
		\midrule
		\rowcolor{gray!20}
		BMT(pred) & - & \uline{45.06} &  \uline{22.40} &  157.25 &  5.77 \\
		\rowcolor{gray!20}
		BMT(gt) & - & \textbf{47.58} & \textbf{22.60} & \textbf{137.20} & \textbf{3.48} \\
		\bottomrule
	\end{tabular}
	\\[3pt]
	\footnotesize $\uparrow$ indicates higher values are better. Best values are in bold. Suboptimal results are underlined.
	\label{table1}
\end{table}

\begin{table*}[htbp]
	\centering
	\caption{Quantitative segmentation evaluation of the images generated by the different methods on the WHU-OPT-SAR dataset, obtained with a common FCN-32s evaluator.}
	\label{table2}
	\setlength{\tabcolsep}{10pt}
	\begin{tabular}{lccc|ccccccc}
		\toprule
		\multicolumn{1}{c}{\multirow{2}{*}{\textbf{Method}}} & \multicolumn{1}{c}{\multirow{2}{*}{\textbf{Year}}} & \multicolumn{2}{c}{\textbf{Results}} & \multicolumn{7}{c}{\textbf{Category(mIoU)}} \\
		\cmidrule(lr){3-4} \cmidrule(lr){5-11}
		& & \textbf{mIoU} \(\uparrow\)& \textbf{mPA} \(\uparrow\)& \textbf{Farmland} & \textbf{City} & \textbf{Village} & \textbf{Water} & \textbf{Forest} & \textbf{Road} & \textbf{Others} \\
		\midrule
		CycleGAN  & 2017  & 25.89 & 35.05 & 34.53 & \uline{41.06} & 8.62  & 47.09 & 48.50 & 0.62  & 0.86  \\
		Pix2pix   & 2017  & 22.41 & 30.46 & 38.29 & 14.97 & 8.05  & 35.61 & 58.92 & 0.33  & 0.67  \\
		Pix2pixHD & 2018  & 27.20 & 36.33 & 39.88 & 24.47 & \uline{14.78} & 47.37 & 60.98 & 1.22  & 1.69  \\
		StegoGAN  & 2024  & 28.20 & 37.53 & 37.30 & 35.15 & 11.63 & 56.83 & 54.69 & 1.08  & 0.71  \\
		PatchGCL  & 2024  & 24.95 & 33.61 & 31.58 & 24.51 & 9.84  & 50.66 & 56.32 & 0.08  & 1.67  \\
		S-CycleGAN & 2019 & 22.46 & 30.77 & 25.38 & 22.35 & 10.29 & 40.82 & 56.08 & 0.77  & 1.52  \\
		WFLM-GAN   & 2022 & \uline{29.32} & \uline{38.46} & 40.02 & \textbf{41.58} & \textbf{15.20} & 42.39 & 61.85 & 1.63  & 2.55  \\
		ControlNet   & 2023 & 26.58 & 35.57 & 34.44 & 29.83 & 10.74 & 53.75 & 55.51 & 0.58 & 1.19  \\
		HVTC-GAN   & 2026 & 26.70 & 34.90 & 39.56 & 28.46 & 10.65 & 45.26 & 62.21 & 0.02 & 0.79 \\ 
	
		\midrule
		\rowcolor{gray!20}
		BMT(pred)      & - & 27.71  & 35.71  & \uline{43.68} & 5.76  & 10.77 & \uline{57.75}  & \uline{65.02} & \uline{1.81}  & \uline{9.21}  \\
		\rowcolor{gray!20}
		BMT(gt)  & -  & \textbf{31.97}  & \textbf{40.18}  & \textbf{49.59} & 13.60  & 12.45  & \textbf{60.04}  & \textbf{68.06}  & \textbf{6.03}  & \textbf{14.02}  \\
		\bottomrule
	\end{tabular}
	\\[3pt]
	\footnotesize $\uparrow$ indicates higher values are better. Best values are in bold. Suboptimal results are underlined.
\end{table*}

\begin{table*}[htbp]
	\centering
	
	\caption{Quantitative segmentation comparison with dedicated segmentation models on the WHU-OPT-SAR dataset.}
	\label{table2-2}
	\setlength{\tabcolsep}{10pt}
	\begin{tabular}{lccc|ccccccc}
		\toprule
		\multicolumn{1}{c}{\multirow{2}{*}{\textbf{Method}}} & \multicolumn{1}{c}{\multirow{2}{*}{\textbf{Year}}} & \multicolumn{2}{c}{\textbf{Results}} & \multicolumn{7}{c}{\textbf{Category(mIoU)}} \\
		\cmidrule(lr){3-4} \cmidrule(lr){5-11}
		& & \textbf{mIoU} \(\uparrow\)& \textbf{mPA} \(\uparrow\)& \textbf{Farmland} & \textbf{City} & \textbf{Village} & \textbf{Water} & \textbf{Forest} & \textbf{Road} & \textbf{Others} \\
		\midrule
		FCN    & 2015 & 37.90 & 47.48 & 47.53 & 54.47 & 17.54 & 66.03 & 66.43 &  5.80 &  7.52 \\
		Swin-Unet  & 2022 & 40.96 & 53.33 & 46.96 & 58.88 & 24.93 & 72.57 & 65.94 &  7.70 &  9.76 \\
		DeepLabV3+  & 2018  & 45.42 & 56.78 & 53.47 & 61.31 & 26.37 & 72.37 & 71.28 & 27.40 &  5.72 \\
		U-Net      & 2015  & 46.15 & 57.56 & 54.06 & 58.75 & 26.32 & 74.83 & 71.65 & 26.88 & 10.54 \\
		PSPNet    & 2017    & 46.91 & 57.17 & 54.94 & 63.27 & 26.89 & 74.76 & 72.08 & 28.07 &  8.38 \\
		HRNet      & 2019   & 48.51 & 59.40 & 56.68 & 62.20 & 28.91 & 75.61 & 72.54 & \textbf{31.28} & 12.37 \\
		SegFormer & 2021   & \uline{49.31} & 60.40 & \uline{57.50} & 64.44 & \uline{29.34} & 75.62 & 73.56 & \uline{30.47} & 14.24 \\
		LoG-CAN & 2023 & 48.46 & 59.55 & 56.89 & 64.95 & 27.71 & 75.37 & \uline{73.77} & 26.30 & 14.25 \\
		SFFNet & 2024 &  44.61 & 56.09 & 53.67 & 61.89 & 26.86 & 74.15 & 70.30 & 12.50 & 12.89 \\
		DOCNet & 2024 &  \textbf{49.82} & \uline{60.50} & \textbf{58.68} & \textbf{66.10} & \textbf{29.49} & \textbf{76.15} & \textbf{74.50} & 27.65 & \textbf{16.20}\\
		AFENet  &2025 & 48.96 & \textbf{60.70}  & 57.18 & \uline{65.76} & 28.78 & \uline{75.77} & 72.63 & 27.10 & \uline{15.46}  \\ 
		DeepKANSeg  &2026 & 44.75 & 54.69  & 54.64 & 64.56 & 26.62 & 74.24 & 72.67 & 16.83 &  3.70   \\ 
		HVTC-GAN   & 2026   & 37.79 & 47.42 & 47.73 & 37.66 & 19.35 & 68.44 & 67.33 & 12.65 & 11.40 \\
		\midrule
		\rowcolor{gray!20}
				BMT     & -  & 44.40 & 55.42 & 53.19 & 60.24 & 25.66 & 72.98 & 71.17& 13.28 & 14.30 \\
		\bottomrule
	\end{tabular}
	\\[3pt]
	\footnotesize $\uparrow$ indicates higher values are better. Best values are in bold. Suboptimal results are underlined.
\end{table*}

	Table~\ref{table1} quantifies that inspection half. The deployable setting is BMT(pred), in which the generator is conditioned on the model's own segmentation. It leads the compared translators on SSIM and PSNR. ControlNet is closer on FID and KID, so distribution matching and pixel-wise agreement do not rank the same way; HVTC-GAN remains the strongest unconditioned GAN on SSIM and PSNR. Ground-truth maps were used to teacher-force generation during training; they are not used here. The gain is a closer match to the optical domain in structure and texture---the appearance an operator uses when Gaofen-1 is unavailable---not a claim that the synthesis recovers optical reflectance. BMT(gt) is higher still and is reported only as an upper bound: the leftover gap is how much the generator still depends on a map more accurate than its own prediction. The absolute level remains modest because the test tiles do not share a parent scene with training.

	Figure~\ref{fig:seg-whu} and Table~\ref{table2} ask why the inspection half cannot be the interpretation half. A single FCN-32s pre-trained on optical imagery is applied to every translated image, so differences come from the generated pixels rather than from a change of decoder. On water-dominated scenes, BMT reconstructs a connected water body with a relatively complete shoreline; most baselines truncate the water, HVTC-GAN being the main exception. On road and built-up scenes BMT is smoother, at the cost of roads that are sometimes drawn too wide. Quantitatively, every generation pipeline lands in a low band, with mIoU in the low twenties to low thirties. A photometric translator therefore need not preserve the layout that a downstream optical segmentor requires. BMT(pred) is comparable to the strongest generation baselines and slightly below WFLM-GAN and StegoGAN on this probe, and ControlNet's FID advantage does not carry over here; City IoU under the predicted condition is the clear failure mode, which is why a layout-constrained inspection image still cannot carry the land-cover decision. BMT(gt) becomes the highest row overall, so how a frozen optical FCN reads the image follows the injected map, while remaining well below a dedicated SAR decoder. The probe is seven-class land cover, not a binary foreground task. Its content for the framework is not a ranking of S2O. It is the reason land-cover decisions stay on the radar measurement: a synthesised image is a visualisation, not a substitute sensor, so translation alone cannot carry interpretation even though generation was trained with labels.

	Table~\ref{table2-2} is the interpretation half---labels predicted from SAR, not from the fake optical image---and is the comparison that belongs next to dedicated segmentors. This is the SAR-side product of the same labelled, paired training run: a land-cover map obtained from radar, with the optical reconstruction target used as extra supervision on that split rather than as a way to drop annotations. BMT sits in the middle of that group: above FCN, Swin-Unet and the serial baseline HVTC-GAN, close to SFFNet and DeepKANSeg, within about one point of DeepLabV3+, and below U-Net, PSPNet, HRNet, LoG-CAN, AFENet, SegFormer and DOCNet. DOCNet is the strongest dedicated row. The gap is not uniform. On water and forest, BMT is already in the same band as DeepLabV3+ and U-Net; on Others it is among the higher rows, behind DOCNet and AFENet. What separates HRNet and SegFormer most sharply is Road, a thin minority class for which those two networks keep high-resolution streams or a Mix-FFN decoder; farmland, city and village also trail, but by a smaller margin. BMT's labelling head is the counterpart of the generative decoder, so it inherits the encoder's downsampling. That is a cost of sharing on linear objects, not a collapse of the area classes that dominate land-cover mapping.

	The evidence that the translation objective helps SAR labelling is not the comparison with DOCNet or SegFormer. It is the comparison with the same backbone trained without translation on the same labelled tiles, and with a serial pipeline that never lets labelling revise generation. A segmentation-only run of BMT is lower than the joint model (Table~\ref{ab-cond}); HVTC-GAN is lower still. Optical reconstruction therefore helps SAR labelling as extra supervision on a fully annotated split: it does not mint labels, it does not drop labels, and it does not make training cheaper than a dedicated segmentor. Dedicated networks keep a representation that never has to satisfy an adversarial and a perceptual loss at the same time, which is why they pull ahead on Road; uncertainty weighting reduces that tension but does not remove it. The intended reading of Table~\ref{table1} and Table~\ref{table2-2} together is the value claim of Section~I. Training used paired optics and ground-truth maps; testing uses neither. An unconditional translator does not label the SAR tile. A dedicated segmentor does not emit an optical-like view. BMT returns both from one SAR pass: a layout-constrained inspection readout that leads the compared translators on SSIM and PSNR under the predicted condition, and a SAR-side map that is near DeepLabV3+ on water and forest, trails the strongest single-task models on thin objects, and is stronger than the same backbone trained on radar labels and speckle alone.

\subsection{Ablation Studies}

The collaborative backbone combines the LocalViTBlock, the condition injection mechanism, and several loss terms. To assess the contribution of each component, we conduct the ablation experiments reported below on the WHU-OPT-SAR dataset.

\begin{table*}[htbp]
	\centering
	\caption{Ablation on WHU-OPT-SAR under the predicted-condition protocol.}
	\label{ab-cond}
	\renewcommand{\arraystretch}{1.0}
	\setlength{\tabcolsep}{8pt}
	\begin{tabular}{lcccccc}
		\toprule
		\textbf{Configuration} & \textbf{SSIM} \(\uparrow\)(\%) & \textbf{PSNR} \(\uparrow\)(dB) & \textbf{FID} \(\downarrow\) & \textbf{KID} \(\downarrow\)(\(\times 100\)) & \textbf{mIoU} \(\uparrow\) & \textbf{mPA} \(\uparrow\) \\
		\midrule
		\multicolumn{7}{l}{\textit{Task branches (the joint model is the default row below)}} \\
		generation branch only & 43.19 & 21.50 & 162.05 & 6.70 & -- & -- \\
		segmentation branch only & -- & -- & -- & -- & 40.57 & 52.22 \\
		\midrule
		\multicolumn{7}{l}{\textit{Condition injection}} \\
		w/o conditioning\(^{\dagger}\) & \uline{46.37} & \uline{22.71} & 162.82 & 6.80 & 43.40 & 54.28 \\
		+ bottleneck only & \textbf{46.61} & 22.63 & 161.70 & 6.10 & 43.99 & 54.89 \\
		+ shallow levels \(\mathbf{f}_0,\mathbf{f}_1\) & 45.37 & 22.05 & 159.22 & 5.53 & 43.88 & 54.68 \\
		+ deep levels \(\mathbf{f}_2,\mathbf{z}\) & 46.10 & 22.24 & 157.20 & 6.12 & 43.70 & 54.73 \\
		\rowcolor{gray!20}
		+ all levels (default, joint model) & 45.06 & 22.40 & 157.25 & 5.77 & \textbf{44.40} & \textbf{55.42} \\
		\midrule
		\multicolumn{7}{l}{\textit{Output head (the default head is the row above)}} \\
		plain refinement + to RGB & 43.20 & 21.81 & \textbf{152.02} & \textbf{5.44} & \uline{44.20} & \uline{55.34} \\
		w/o anti-aliasing & 45.76 & 22.16 & 155.00 & 5.67 & 44.20 & 54.79 \\
		w/o color correction & 44.30 & 21.93 & 155.03 & \uline{5.45} & 43.79 & 55.15 \\
		\midrule
		\multicolumn{7}{l}{\textit{Injection initialization (the default is zero-initialized)}} \\
		standard initialization instead of zero & 44.34 & 21.97 & \uline{152.60} & 5.64 & 43.57 & 54.22 \\
		\midrule
		\multicolumn{7}{l}{\textit{LocalViTBlock}} \\
		w/o LocalViTBlock & 45.35 & 21.91 & 162.49 & 6.03 & 42.87 & 54.92 \\
		w/o LocalViTBlock and w/o conditioning\(^{\dagger}\) & 46.25 & \textbf{22.72} & 172.73 & 7.73 & 42.81 & 53.32 \\
		\bottomrule
	\end{tabular}
	\\[3pt]
	\footnotesize \(\uparrow\) / \(\downarrow\) indicate whether higher or lower values are better. Best values are in bold. Suboptimal results are underlined. \(^{\dagger}\)No condition; generation metrics are reported without a semantic input.
\end{table*}

	\begin{table*}[htbp]
	\centering
	\caption{Ablation on WHU-OPT-SAR with the generation pass conditioned on the ground-truth label map.}
	\label{ab-cond-ub}
	\renewcommand{\arraystretch}{1.0}
	\setlength{\tabcolsep}{14pt}
	\footnotesize
	\begin{tabular}{lcccc}
		\toprule
		\textbf{Configuration} & \textbf{SSIM} \(\uparrow\)(\%) & \textbf{PSNR} \(\uparrow\)(dB) & \textbf{FID} \(\downarrow\) & \textbf{KID} \(\downarrow\)(\(\times 100\)) \\
		\midrule
		\multicolumn{5}{l}{\textit{Condition injection}} \\
		+ bottleneck only & 46.54 & \textbf{22.77} & 158.84 & 6.02 \\
		+ shallow levels & 47.77 & 22.73 & 139.05 & 3.64 \\
		+ deep levels & 46.79 & \uline{22.64} & 149.38 & 5.12 \\
		\rowcolor{gray!20}
		+ all levels (default) & 47.58 & 22.60 & 137.20 & 3.48 \\
		\midrule
		\multicolumn{5}{l}{\textit{Output head (the default head is the row above)}} \\
		plain refinement + to RGB & 45.30 & 22.45 & \uline{135.05} & \uline{3.47} \\
		w/o anti-aliasing & \textbf{47.99} & \textbf{22.77} & 136.74 & 3.79 \\
		w/o color correction & 46.62 & 22.57 & \textbf{133.86} & \textbf{3.36} \\
		\midrule
		\multicolumn{5}{l}{\textit{LocalViTBlock}} \\
		w/o LocalViTBlock & \uline{47.93} & 22.62 & 136.97 & 3.62 \\
		\midrule
		\multicolumn{5}{l}{\textit{Injection initialization (the default is zero-initialized)}} \\
		standard initialization instead of zero & 46.5 & 22.58 & 136.81 & 4.01 \\
		\bottomrule
		
	\end{tabular}
	\\[3pt]
	\footnotesize \(\uparrow\) / \(\downarrow\) indicate whether higher or lower values are better. Best values are in bold. Suboptimal results are underlined.
	\end{table*}

We first compare the single-task branches with the joint model in Table~\ref{ab-cond}. Sharing the encoder raises segmentation well above the segmentation-only run on the same labelled tiles. Generation is not a like-for-like comparison: the generation branch trained alone has no segmentation head and is therefore unconditional. Against that run, the joint model without a condition is better on SSIM and PSNR and essentially tied on FID and KID. Feeding it its own predicted map then improves FID and KID, while SSIM and PSNR remain higher without a condition. Joint training therefore helps the segmentor without falling behind the generation-only branch on pixel-wise scores; the predicted condition is what brings the translated image closer to the optical domain in distribution, not in SSIM.

\begin{figure}[htbp]
	\centering
	\includegraphics[width=0.8\columnwidth]{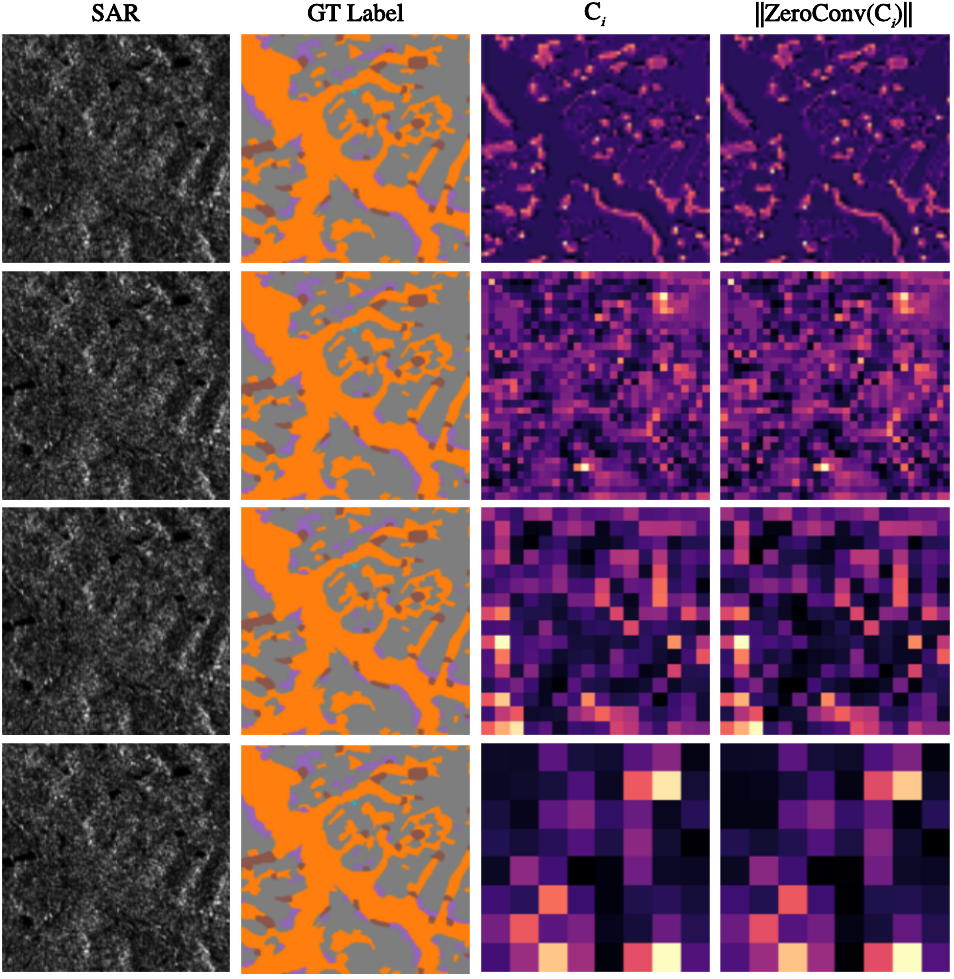}
	\caption{Conditioning path of the default model. Columns: SAR patch, training-time semantic map, condition features $\mathbf{c}_0$--$\mathbf{c}_3$, and the residual added at the matching encoder level. Rows: encoder stages from fine (top) to coarse (bottom).}
	\label{fig:cond-path}
\end{figure}

\begin{figure}[htbp]
	\centering
	\includegraphics[width=0.8\columnwidth]{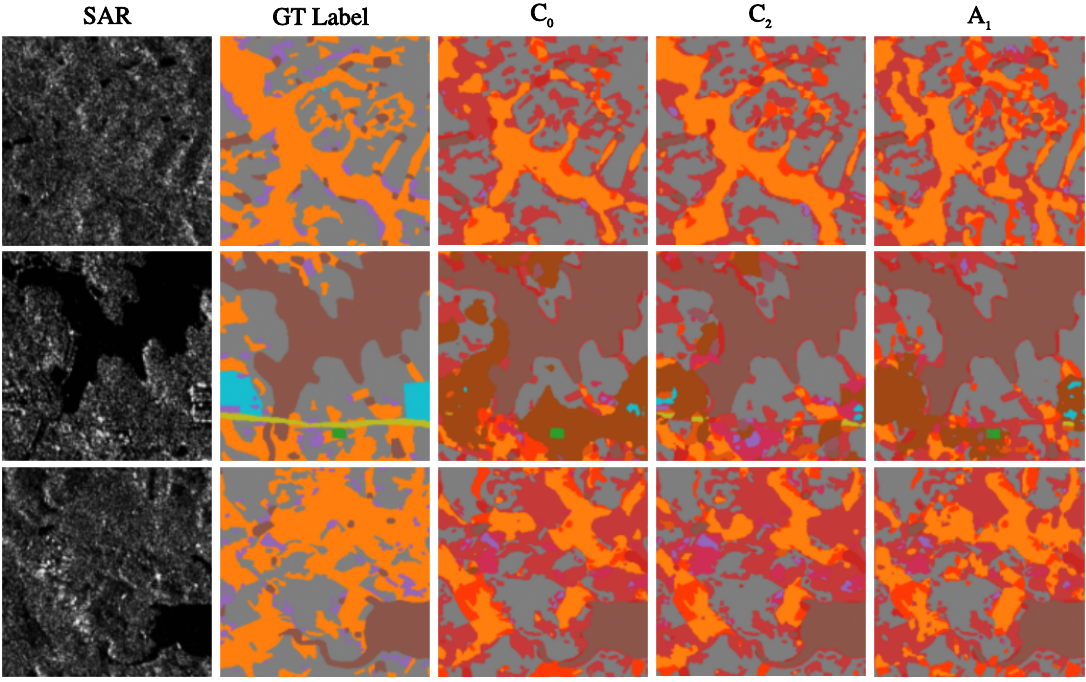}
	\caption{Segmentation maps of three ablation variants. Rows: test scenes. Columns: SAR patch, ground-truth labels, and predictions without conditioning, with the default model, and without the LocalViTBlock. Misclassified pixels are shown in red.}
	\label{fig:seg-ablation}
\end{figure}

\begin{figure}[htbp]
	\centering
	\includegraphics[width=0.8\columnwidth]{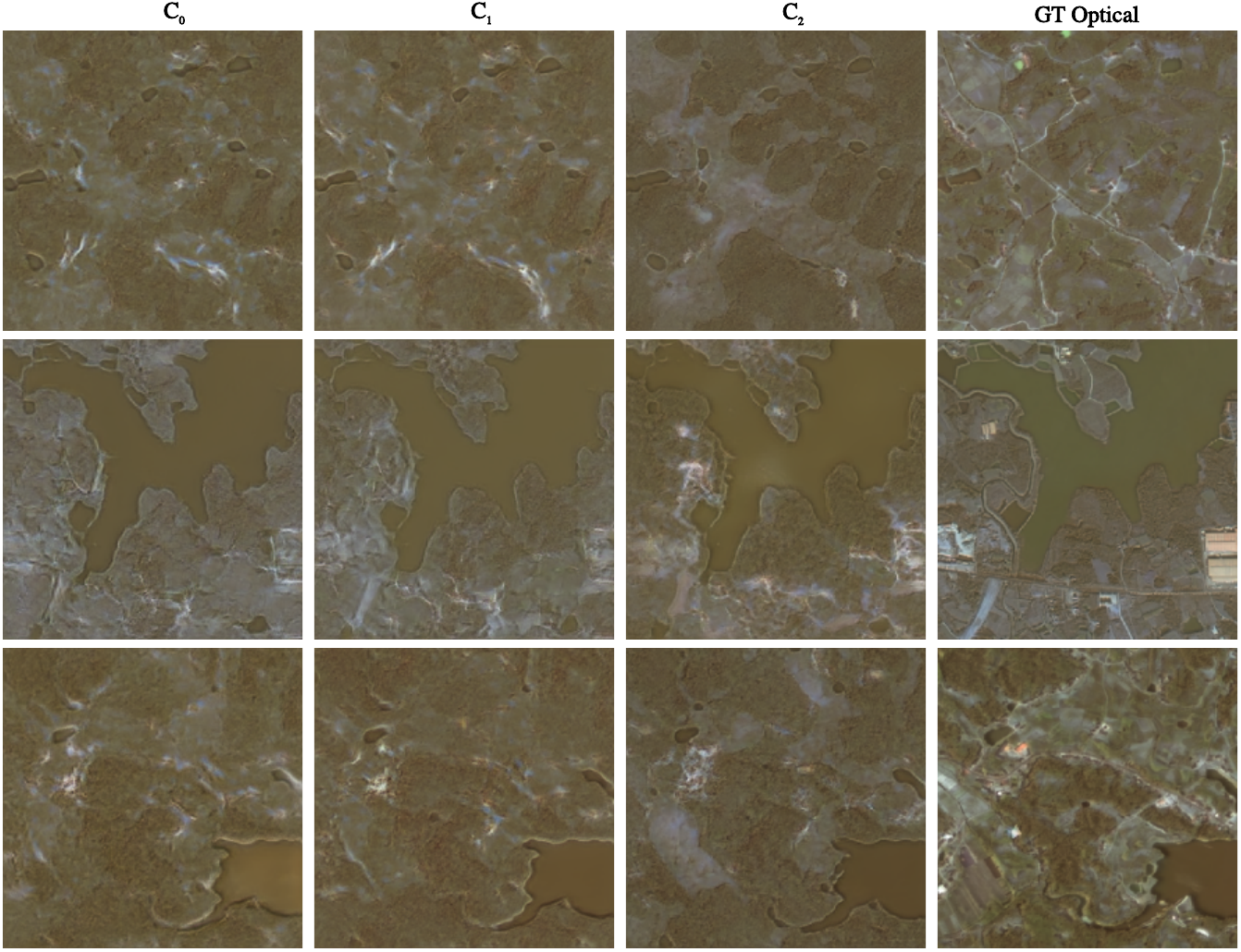}
	\caption{Translated images under different injection layouts (predicted condition). Rows: water, city and road scenes. Columns: no conditioning, bottleneck-only injection, all-level injection, and the optical reference.}
	\label{fig:inj-layout}
\end{figure}

Tables~\ref{ab-cond} and~\ref{ab-cond-ub} separate having a semantic condition from where it is injected. Adding the condition improves the shared encoder's mIoU and the distribution-level generation scores; Figure~\ref{fig:cond-path} shows the pyramid and the residual at each encoder level. The injection depth does not admit a single winner under the predicted-condition protocol used at test time. Bottleneck-only is best on SSIM, all-level injection is best on mIoU, and deep and all-level injection are essentially tied on FID; Figure~\ref{fig:inj-layout} is correspondingly close in appearance. We keep all-level injection as the default because it is the best labelling configuration and competitive on FID, not because every extra depth is independently necessary. The picture changes when the same layouts receive a ground-truth map: all-level injection then leads clearly on FID and KID, which is what one expects of a spatial prior that is accurate at every scale. Adversarial training still varies from run to run, so we do not rank near-tied rows. The LocalViTBlock is more consistent than the layout choice. Removing it worsens FID, KID and mIoU with or without a condition, while SSIM barely moves and PSNR drops. Figure~\ref{fig:seg-ablation} shows more misclassified pixels, in line with the mIoU drop.

\begin{figure}[htbp]
	\centering
	\includegraphics[width=\linewidth]{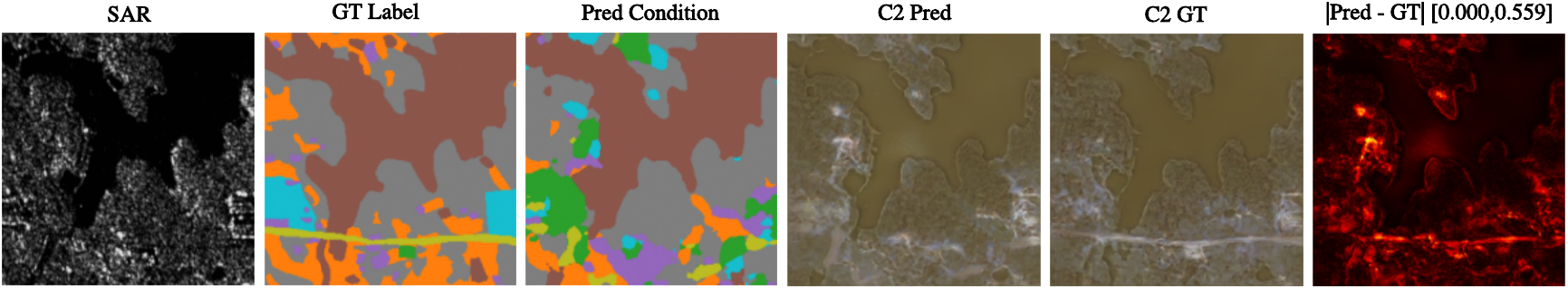}
	\caption{Predicted- versus ground-truth-condition generation on a city scene. Columns: SAR patch, ground-truth labels, image with the predicted condition, image with the ground-truth condition, their absolute difference, and the condition used at test time.}
	\label{fig:cond-diagnostic}
\end{figure}

	Color correction, anti-aliasing, the uncertainty weights of Table~\ref{tab:weighting}, and zero versus standard initialization of the injection convolutions do not isolate as cleanly. Segmentation metrics, and in most cases SSIM and PSNR, favour the default; removing anti-aliasing is the exception on SSIM, while mIoU still drops. FID and KID often move the other way by about as much as run-to-run scatter. We keep those defaults as a package that prefers pixel-level and labelling accuracy, without assigning unique credit to each piece. Replacing the predicted map by the annotation, by contrast, cuts FID and KID by a much larger margin than any of these architectural variants (Table~\ref{ab-cond-ub}). Most of the remaining gap to the optical distribution is therefore the accuracy of the predicted condition, not a limitation of the generator. Figure~\ref{fig:cond-diagnostic} isolates that gap on one city scene.

\subsection{Computational Efficiency and Limitations }

Table~\ref{tab:efficiency} reports computational cost for representative translators (Pix2pix, WFLM-GAN, HVTC-GAN), dedicated SAR segmentors (DeepLabV3+, SegFormer), and BMT as generation-only, segmentation-only, and the joint predicted-condition model. The comparison is whether sharing the encoder returns both outputs more cheaply than two independent models or a heavy serial translator.

\begin{table}[t]
	\centering
	\caption{Computational cost of representative translators, dedicated segmentors, and BMT. Joint BMT uses the predicted-condition protocol.}
	\label{tab:efficiency}
	\renewcommand{\arraystretch}{1.05}
	\setlength{\tabcolsep}{4pt}
	\begin{tabular}{lccc}
		\toprule
		\textbf{Method} & \textbf{Params (M)} & \textbf{FLOPs (G)} & \textbf{Latency (ms)} \\
		\midrule
		Pix2pix & 57.18& 18.15 & 1.19 \\
		WFLM-GAN & 217.11 & 971.12 & 16.96 \\
		HVTC-GAN & 201.10 & 205.03 & 6.11 \\
		\midrule
		DeepLabV3+ & 54.71 & 60.83 & 6.48 \\
		SegFormer & 84.60 & 24.20 & 27.85 \\
		\midrule
		BMT (generation only) & 101.58 & 41.67 & 14.93 \\
		BMT (segmentation only) & 86.21 & 29.68 & 11.99 \\
		BMT (joint) & 138.79 & 74.79 & 27.40 \\
		\bottomrule
	\end{tabular}
	\\[3pt]
	\footnotesize Measurements on a single GPU (FP32, batch size 1, $256\times256$). Latency is averaged after warm-up. Joint BMT runs segmentation first and conditions generation on the predicted map.
\end{table}

Joint training does not remove the tension between heads: FID still trades against pixel-wise and labelling accuracy, and the shared decoder remains below the strongest dedicated segmentors. The leftover generation gap is the predicted condition; a frozen optical FCN on the translated image also stays in a low mIoU band. Optical-to-SAR based on scattering and speckle, extra paired sensors, and long-tailed labelling without background-saturated mIoU are left for later work.

\begin{table}[htbp]
	\centering
	\caption{Task-weighting ablation on WHU-OPT-SAR.}
	\label{tab:weighting}
	\renewcommand{\arraystretch}{1.0}
	\setlength{\tabcolsep}{6pt}
	\footnotesize
	\begin{tabular}{lcc}
		\toprule
		\textbf{Metric} & \textbf{Learned uncertainty} & \textbf{Fixed equal weights} \\
		\midrule
		SSIM \(\uparrow\)(\%) & \textbf{45.06} & 43.65 \\
		PSNR \(\uparrow\)(dB) & \textbf{22.40} & 21.17 \\
		FID \(\downarrow\) & 157.25 & \textbf{153.10} \\
		KID \(\downarrow\)(\(\times 100\)) & 5.77 & \textbf{5.26} \\
		mIoU \(\uparrow\) & \textbf{44.40} & 43.32 \\
		mPA \(\uparrow\) & \textbf{55.42} & 54.76 \\
		\bottomrule
	\end{tabular}
	\\[3pt]
	\footnotesize \(\uparrow\) / \(\downarrow\) indicate whether higher or lower values are better. Best values are in bold.
\end{table}

\section{Conclusion}
This paper proposes BMT, a unified multi-task framework that collaboratively optimizes S2O translation and semantic segmentation through a shared hierarchical ViT encoder. Training uses paired optical images and full pixel labels, including ground-truth maps to teacher-force generation; testing uses SAR alone. The model then supplies an inspectable optical-like readout and a SAR-side land-cover map regularized by that appearance cue. On WHU-OPT-SAR, the predicted-condition model attains the best SSIM and PSNR among the compared translation methods and improves semantic segmentation over the same backbone trained for labelling alone on the same split.

\bibliographystyle{unsrt}
\bibliography{reference}
	
\end{document}